\documentclass{article}

\usepackage{spconf,amsmath,graphicx}
\usepackage{booktabs}
\usepackage{enumitem}
\usepackage{array}
\usepackage{multirow}
\usepackage{diagbox}
\usepackage{color}

\usepackage{fancyhdr}

\pdfoutput=1

\usepackage[pagebackref=true,breaklinks=true,letterpaper=true,colorlinks,bookmarks=false]{hyperref}

\let\oldfootnote\footnote
\def\footnote{\ifhmode\unskip\fi\oldfootnote}

\newcommand{\etal}{\textit{et al}.}

\newlength{\tempdima}
\newcommand{\rowname}[1]
{\rotatebox{90}{\makebox[\tempdima][c]{{\scriptsize #1}}}}

\title{VISUAL-QUALITY-DRIVEN LEARNING FOR UNDERWATER VISION ENHANCEMENT}

\name{Walysson V. Barbosa, Henrique G. B. Amaral, Thiago L. Rocha, Erickson R. Nascimento}

\address{Universidade Federal de Minas Gerais (UFMG), Brazil \\
	\{wallyb, erickson\}@dcc.ufmg.br, \{henriquegrandinetti, thiagolagesrocha\}@gmail.com}

\begin{document}
	
\maketitle

\thispagestyle{fancy}
\fancyhf{}
\fancyfoot[L]{978-1-4799-7061-2/18/\$31.00~\copyright2018~IEEE}
\chead{In Proceedings of the 2018 IEEE International Conference on Image Processing (ICIP) \\ The final publication is available at: https://doi.org/10.1109/ICIP.2018.8451356}
\setlength{\headsep}{0.08 in}

\begin{abstract}
	
	The image processing community has witnessed remarkable advances in enhancing and restoring images. Nevertheless, restoring the visual quality of underwater images remains a great challenge. End-to-end frameworks might fail to enhance the visual quality of underwater images since in several scenarios it is not feasible to provide the ground truth of the scene radiance. In this work, we propose a CNN-based approach that does not require ground truth data since it uses a set of image quality metrics to guide the restoration learning process.  The experiments showed that our method improved the visual quality of underwater images preserving their edges and also performed well considering the UCIQE metric.
	
\end{abstract}


\section{Introduction} \label{sec:intro}

With image processing and learning approaches rapidly evolving, the ability to make sense of what is going on in a single picture is improving in both scalability and accuracy. However, restoring the visual quality of images acquired from participating media such as underwater environments remains a significant challenge for most of the image processing techniques. After all, underwater images are crucial in many critical applications, such as biological research, maintenance of marine vessels, and studies of submerged archaeological sites that cannot be removed from the water.

Despite remarkable advances in restoring underwater images with learning methods like Convolutional Neural Networks (CNN), the number of images and the quality of the ground truth data used in training limits these methods. In underwater environments, the light is scattered and absorbed when traveling its way to the camera. As a consequence, objects distant from the camera appear dimmer, with low contrast and color distortion. The ground truth of an underwater image is then another image of the same scene but immersed in a non-participating media without scattering and absorption. Building datasets with high quality and a large number of images is hard or infeasible, in most cases it is difficult to acquire images of an underwater scene in a non-participating media, \emph{e.g.}, images taken from under the sea. Hence, the ability to work with a small number of images or with simulated underwater images plays a key role in restoring the visual quality of underwater images.

In this work, we propose a new learning approach for restoring the visual quality of underwater images. Our approach aims at restoring the images by working with simulation data and not demanding a large amount of real data. A set of image quality metrics guides the optimization process toward the restored image. The experiments showed that our approach outperforms other methods qualitatively and quantitatively when considering the quality metric UCIQE~\cite{yang:2015}.

\paragraph*{Related Work.} Early works on image restoration relied on image processing techniques, which focused mostly on enhancing the contrast level of the scene~\cite{Bazeille2006, gu:2016, zheng:2016}. 

In the past decade, the methods based on physical models have emerged as effective approaches to predict the original scene radiance~\cite{Trucco2006,SchechnerCVPR2007,Nascimento2009,he:2011,drews:2016,peng:2017}. A typical approach is the Dark Channel Prior (DCP)~\cite{he:2011}. The DCP calculation takes the minimum value per channel at each pixel of an image. Mostly applied to outdoor haze-free images, the idea is that at least one of the intensity values from all color channels tends to zero. Because the assumption of the dark channel might not hold in underwater scenes, Drews~\etal~\cite{drews:2016} presented the Underwater Dark Channel Prior (UDCP). The authors used only the blue and green channels since the red channel is dramatically absorbed in underwater. The UDCP achieved better results than those obtained by using the DCP.

\begin{figure*}[t!]
	\centering
	\includegraphics[width=.85\textwidth]{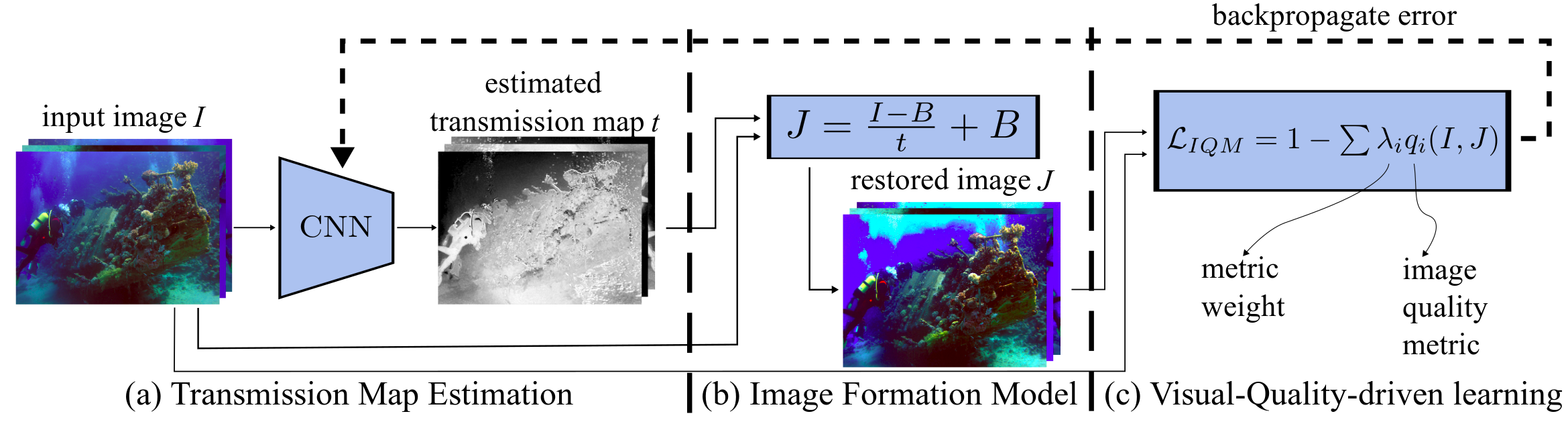}
	\caption{Diagram of our two-stage learning. First, we fine-tune the CNN using ground-truth transmission maps, applying the mean squared error loss in the training process. Second, we take the model and adapt it by including the image restoration process (b). Finally, we perform new training in the network minimizing a loss function based on image quality metrics.}
	\label{fig:methodology}
\end{figure*}

More recently, learning techniques have shown promising results when used for enhancing the visual quality of images taken from participating media~\cite{cai:2016, ren:2016, li:2017, liu:2017}. Li~\etal~\cite{li:2017} proposed a method to generate synthetic underwater images from regular air ones. They used a generative adversarial network (GAN) to simulate the water attenuation and perform the image transformation. The major drawback of their method is the requirement of a large dataset covering many different situations for a useful generalization. Ren~\etal~\cite{ren:2016} estimate the transmission map using two convolutional networks. While one network extracts a general, but a rough version of the map, the second one performs local improvements. Cai~\etal~\cite{cai:2016} presented a network for terrestrial images restoration, the DehazeNet. Their goal is to remove the haze by minimizing the error between expected and estimated transmission maps, which they state it is the key to recover a clean scene. Applying their approach in out of water images achieve impressive results but do not generalize to underwater scenes.

Unlike the methods mentioned above, our approach does not rely on a large dataset. Our premise is based and assessed by image quality metrics. This assumption allows us to obtain results that are directly related to the human sense of quality considering the conditions of underwater images.

\section{Methodology} \label{sec:methodology}

Our methodology comprises two-phase learning. Initially, we perform a supervised training by fine-tuning the DehazeNet~\cite{cai:2016} to estimate the transmission map. Then, we restore the input image according to an image formation model. In the second phase, we minimize a loss function composed of quality metrics to perform image restoration finally. Figure~\ref{fig:methodology} illustrates the whole process.

\paragraph*{Image Formation Model.} In the underwater environment, the image is a combination of the light coming directly from the objects composing the scene and light that was redirected towards the camera. In this work, we use the most commonly referenced image formation model~\cite{fattal:2008, he:2011, drews:2016}:
\begin{equation} \label{eq:uifm}
I(x) = J(x)t(x) + B(1 - t(x)),
\end{equation}
\noindent where $I$ is the observed light intensity,  $J$ the scene radiance,  $B$ the background light, and $t(x) = e^{-\beta d(x)}$ is the transmission map.

The transmission map $t$ gives the amount of light not attenuated, due to scattering or absorption, on a given point $x$ at a distance $d(x)$. The parameter $\beta$ represents the medium attenuation coefficient. Thus, $J$ can be estimated by reformulating Equation~\ref{eq:uifm} as
\begin{equation}\label{eq:j}
J(x) = (I(x) - B)t(x)^{-1} + B.
\end{equation}

Thereby, we only need to estimate $t$ and $B$ to restore an image in the underwater environment. The transmission map $t$ can be estimated using the DehazeNet. Following the prior of Drews~\etal~\cite{he:2011}, we can roughly estimate the background light as being the pixel in the degraded image $I$ whose transmission map value is the highest, limited by a constant $t_0$, \emph{i.e.}, ${B = \max_{y\in\{x|t(x)\leq t_0\}} I(y)}$. The value $t_0$ is chosen as the $0.1\%$ highest pixel.

\paragraph*{Transmission map estimation.}

To adjust the network to our purpose, we proposed to perform supervised training following the guidelines of Cai~\etal~\cite{cai:2016}. It consists in using underwater images as input to the network and comparing its estimated transmission map to the ground-truth maps. The loss function used in this stage is the mean squared error (MSE), defined as
\begin{equation}
\mathcal{L}_{t}(I, J) = \frac{1}{n}\sum_{x=1}^n \left(t_I(x) - t_J(x)\right)^2,
\end{equation}
\noindent where $n$ is the number of pixels, $t_J$ the estimated transmission map and $t_I$ the ground-truth.

\paragraph*{Visual-quality-driven learning.} 

To overcome the absence of ground truth data for underwater scenes, we present an approach that assesses the result by computing a set of Image Quality Metrics (IQM). The IQM set $\mathcal{X}$ yields a multi-objective function that measures the enhancement of four features in the restored image in comparison to the input image. The multi-objective function is given by
\begin{equation} \label{eq:iqm}
IQM(I, J) = \sum_{X \in \mathcal{X}}\lambda_Xq_X(I, J),
\end{equation}
\noindent where $\lambda_X$ is the weight for a feature gain $q_X$.

We choose four metrics that are well correlated to the human visual perception to compose our IQM set: contrast, acutance, border integrity, and gray world prior.

\begin{itemize}[leftmargin=*]
	
	\item \textit{Contrast level:} Underwater images tend to have low contrast as the amount of water between objects and the camera increases. We compute the contrast gain of a restored image $J$ over the degraded $I$ as
	\begin{equation}
	q_C(I, J) = \frac{1}{n}\sum_{x=1}^n \left(C(J, x)^2 - C(I, x)^2\right),
	\end{equation}
	\noindent where $C(I, x)$ is the contrast of the pixel $x$ in the gray image $I$~\cite{Nascimento2009}.
	
	\item \textit{Acutance:} The restoration process should also enhance the acutance metric, which measures the human perception of sharpness. The restoration gain for acutance is given by 
	\begin{equation}
	q_A(I, J) = A(J) - A(I),
	\end{equation}

	\noindent where $A(I) = \frac{1}{n}\sum_{x=1}^n G(I, x)$
	is the acutance of an image $I$. $G(I, x)$ is the gradient strength computed by applying the Sobel operator of $I$ at pixel $x$. 
	
	\item \textit{Border integrity:} It measures the visibility of the borders after the restoration. This measurement allows us to check how much the border increased in regions that were likely to have borders, avoiding random noise to appear in the restored image. It is calculated by
	\begin{equation} \label{eq:bi}
	q_{BI}(I, J) = \frac{\sum_{x=1}^n \left(E(J, x) \times E_d(I, x) \right)}{\sum_{x=1}^n E_d(I, x)},
	\end{equation}
	\noindent where $E$ is an edge detector and $E_d$ is the border image dilated by $5$ pixels. 
	
	\item \textit{Gray world prior:} The fourth feature is the gray world prior~\cite{buchsbaum:1980}, a hypothesis that, under natural circumstances, the mean color of an image tends to gray. Therefore, we evaluate how distant from the gray world our restored image is by computing
	\begin{equation}
	q_G(J) = (I_{max}-I_{min}) - \frac{2}{n}\sum_{x=1}^n (I(x) - I_{m})^2,
	\end{equation}
	\noindent where $I_{max}$ and $I_{min}$ are the maximum and minimum intensity a pixel can have and $I_m = (I_{min} + I_{max})/2$. The first term of $q_G$ makes the metric higher as the distance from the gray world is smaller.
\end{itemize}

After computing the gain in each quality metric, we minimize the subsequent IQM loss:
\begin{equation}
\mathcal{L}_{IQM} = 1 - IQM(I, J).
\end{equation}

The weights of the network are updated by propagating the IQM error backward. Note that this step does not require labeled data. Instead of using ground truth data, our method uses the quality metrics to guide the optimization process for refining the transmission map. As a result, the final image presents a physically-plausible restoration of the input with better quality than the results achieved by other approaches.

\section{Experiments} \label{sec:exp}

\begin{figure*}[t!]
	\centering
	
	\begin{tabular}
		{>{\centering\bfseries}m{0.1ex} >{\centering}m{12ex} >{\centering}m{12ex} >{\centering}m{12ex} >{\centering}m{12ex} >{\centering}m{12ex} >{\centering\arraybackslash}m{12ex}}
		& \hfill {\scriptsize \textbf{Original}} \hfill & \hfill {\scriptsize \textbf{He et al.}} \hfill & \hfill {\scriptsize \textbf{Drews et al.}} & \hfill {\scriptsize \textbf{Tarel et al.}} & \hfill {\scriptsize \textbf{Ancuti et al.}} & \hfill {\scriptsize \textbf{Ours}} \hfill \\
		\vfill \rowname{ancuti1} \vfill &
		\includegraphics[width=0.13\textwidth]{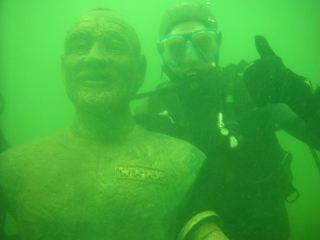} &
		\includegraphics[width=0.13\textwidth]{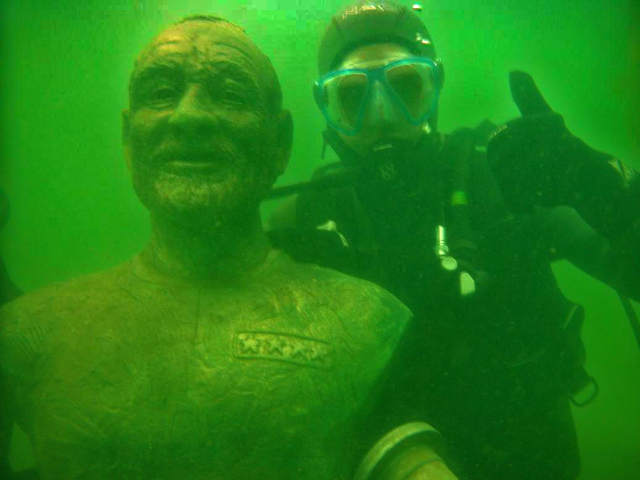} &
		\includegraphics[width=0.13\textwidth]{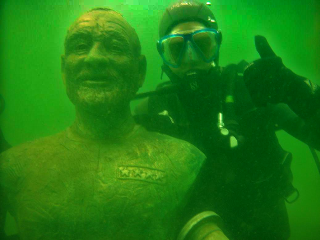} &
		\includegraphics[width=0.13\textwidth]{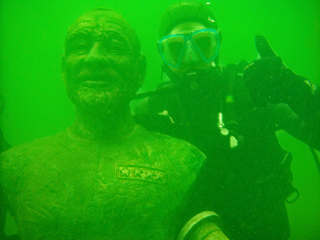} &
		\includegraphics[width=0.13\textwidth]{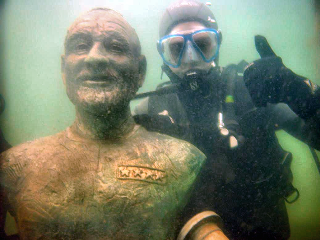} &
		\includegraphics[width=0.13\textwidth]{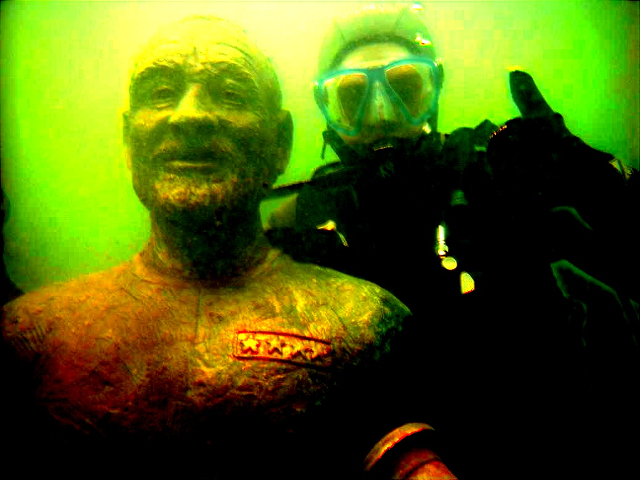} \\
		\vfill \rowname{ancuti2} \vfill &
		\includegraphics[width=0.13\textwidth]{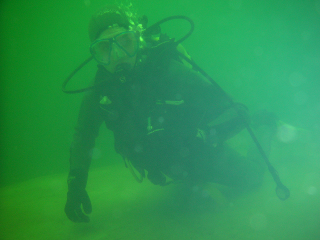} &		
		\includegraphics[width=0.13\textwidth]{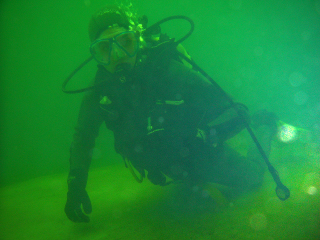} &
		\includegraphics[width=0.13\textwidth]{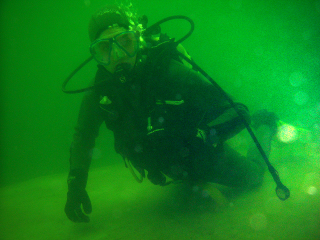} &
		\includegraphics[width=0.13\textwidth]{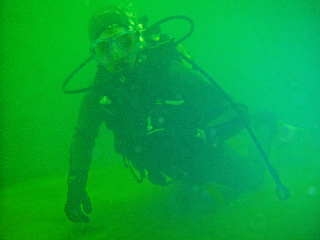} &
		\includegraphics[width=0.13\textwidth]{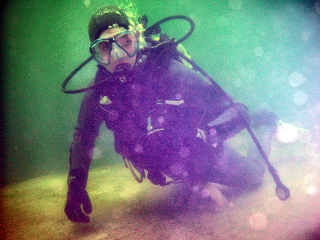} &
		\includegraphics[width=0.13\textwidth]{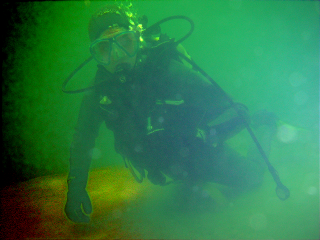} \\
		\vfill \rowname{ancuti3} \vfill &
		\includegraphics[width=0.13\textwidth]{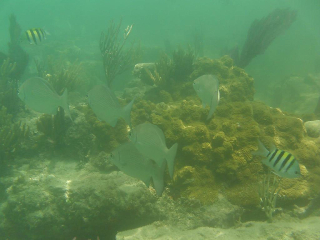} &		
		\includegraphics[width=0.13\textwidth]{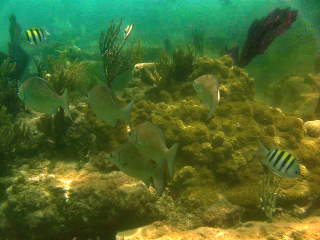} &
		\includegraphics[width=0.13\textwidth]{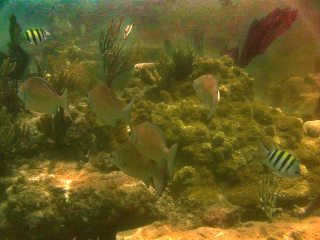} &
		\includegraphics[width=0.13\textwidth]{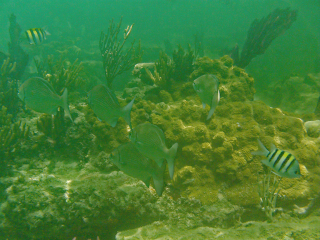} &
		\includegraphics[width=0.13\textwidth]{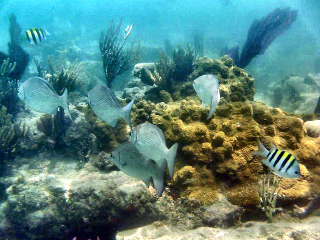} &
		\includegraphics[width=0.13\textwidth]{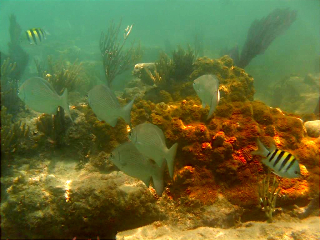} \\
		\vfill \rowname{galdran} \vfill &
		\includegraphics[width=0.13\textwidth]{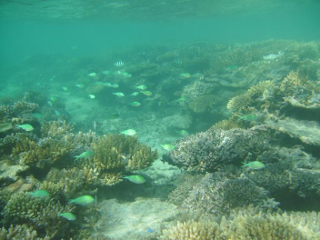} &
		\includegraphics[width=0.13\textwidth]{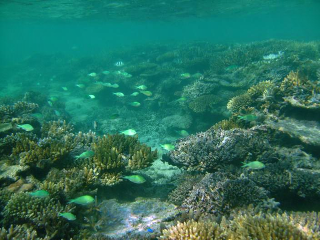} &
		\includegraphics[width=0.13\textwidth]{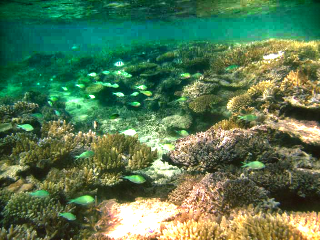} &	
		\includegraphics[width=0.13\textwidth]{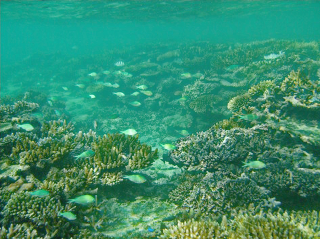} &
		\includegraphics[width=0.13\textwidth]{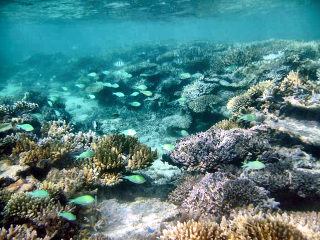} &
		\includegraphics[width=0.13\textwidth]{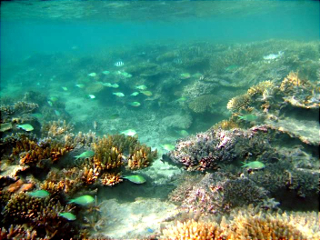} \\
	\end{tabular}
	\caption{Four examples of images from Ancuti et al.'s dataset restored by five different approaches, including ours.}
	\label{fig:results}
\end{figure*}

\paragraph*{Implementation.} Our convolution network has three convolutional layers: $16$ $5{\times}5$ filters in conv1, $16$ $3{\times}3$, $5{\times}5$ and $7{\times}7$ filters in conv2, and a $6{\times}6$ filter in conv3. An element-wise maximum operation is applied to the outputs of the first layer, producing four feature maps which are concatenated and fed to the second convolutional layer. We use a max-pooling in the concatenated outputs of the second layer and a ReLU function in the last layer output. The images are normalized for values between $[0{,}1]$ after the restoration. We restricted $\sum\lambda_X = 1$ to make the IQM score to lie in between $[0{,}1]$, where $1$ stands for the best restoration and $0$ for the worst. The values of lambdas were defined empirically, being $\lambda_C = 0.25$, $\lambda_A = 0.45$, $\lambda_E = 0.05$ and $\lambda_G = 0.25$. The weights of the network are initialized with the weights provided by Cai~\etal~\cite{cai:2016}. In the initial training, we used $300$ $256{\times}256$ images from our synthetic dataset to fine-tune the network. After $1{,}500$ epochs, we switched the training from supervised to unsupervised mode and used $200$ real-scene images. We used $1{,}180$ epochs for the unsupervised training. Experimental data and source code will be publicly available.

\paragraph*{Datasets.} For the supervised phase of our method, we create a set of synthetic underwater images using the physically based rendering engine PBRT~\footnote{https://github.com/mmp/pbrt-v3}. We render two 3D scenes and generate a total of $642$ images positioning the camera in different locations. We set the absorption and scattering coefficients according to Mobley~\cite{mobley:1994}. In the unsupervised phase, we used three datasets. The first dataset is composed of $40$ images  from the underwater-related scene categories of the SUN dataset~\footnote{http://groups.csail.mit.edu/vision/SUN/} and $60$ images from Nascimento~\etal~\cite{Nascimento2009}. In the second dataset, we used a water tank of $126$cm$\times189$cm$\times42$cm dimension with $665$ liters and $3$ configuration of green-tea solutions (pure water, $80$g, and $160$g of tea) to simulate distinct levels of turbidity. We selected $70$ images from a set of $695$ with the camera in different positions and different turbidity levels. The third dataset comprises a subset of images often used by the community and available in the work of Ancuti~\etal~\cite{ancuti:2012}. Although we did not use this third dataset in training, we used it to compare our approach against other methods. 

\paragraph*{Baselines and evaluation metric.} We compared our results against four different techniques: i) DCP, which estimates the depth map of the scene by taking the dark channel; ii) UDCP, a DCP-like prior that does not take into account the red channel when computing the dark channel; iii) Tarel~\etal~\cite{tarel:2009} that use median filtering to restore the image; and iv) Ancuti~\etal~\cite{ancuti:2012}, which restore the image by executing a multi-scale fusion process combining four estimated weight maps. For comparison, we used the UCIQE metric proposed by Yang and Sowmya~\cite{yang:2015}. Yang and Sowmya showed that the UCIQE is in accordance with the human visual assessment. Statistics of chroma, contrast, and saturation are measures correlative to underwater image quality. The UCIQE value is given by
\begin{equation}
UCIQE=c_1\times\sigma_c+c_2\times con_l+c_3\times\mu_s,
\end{equation}
\noindent where $\sigma_c$ is the standard deviation of chroma, $con_l$ is the contrast of luminance, $\mu_s$ is the average of saturation, and $c_1=0.4680$, $c_2=0.2745$, $c_3=0.2576$.

Another metric used in our evaluation is border integrity (Equation~\ref{eq:bi} in Section~\ref{sec:methodology}). It assumes that the restoration process should not degenerate the most relevant borders of a degraded image. At the same time, it should preserve edges continuity. Thus, such metric tells us the gain in border conservation after a recovering process.

\paragraph*{Results and discussion.} Table~\ref{tab:BIResults} shows the final result of the edge conservation. Our method outperformed all methods (zero value indicates no improvement). Figure~\ref{fig:results} shows some visual comparison between the results achieved by all techniques.  Analyzing the images \textit{ancuti3} and \textit{galdran}, aside from our restoration, all others resulted in images that present a certain blur level across all the edges of the image. These results confirm that our technique tends to preserve the valid edges. It also recovers information not much visible before.

Table~\ref{tab:UCIQEResults} shows the UCIQE scores~\footnote{We ran all the experiments using our own implementation of the UCIQE metric following the guidelines in the authors' paper.}. When comparing the restorations using UCIQE metric, one can see that some methods perform better in different images, but our method is the only that presents the best result for more than one image.

\begin{table}[t!]
	\centering
	\caption{Edges conservation (larger is better).}
	\small
	
	\begin{tabular}{@{}lllllll@{}}
		\toprule
		\textbf{Scene} & \textbf{He} & \textbf{Drews} & \textbf{Ancuti} & \textbf{Ours} \\
		\midrule
		\textbf{\textit{ancuti1}} & 0.0000 & 0.0000 & 0.0003 & \textbf{0.0855} \\
		\textbf{\textit{ancuti2}} & 0.0000 & 0.0000 & 0.0006 & \textbf{0.0192} \\
		\textbf{\textit{ancuti3}} & 0.0000 & 0.0000 & 0.0008 & \textbf{0.0009} \\
		\textbf{\textit{galdran}} & 0.0004 & 0.0011 & 0.0062 & \textbf{0.0280} \\
		\bottomrule
	\end{tabular}
	\label{tab:BIResults}
\end{table}

\begin{table}[t!]
	\centering
	\setlength\tabcolsep{2.5pt} % default value: 6pt
	\caption{Visual quality using UCIQE metric (larger is better).}
	\small
	
	\begin{tabular}{@{}lllllll@{}}
		\toprule
		\textbf{Scene} & \textbf{Original} & \textbf{He} & \textbf{Tarel} & \textbf{Drews} & \textbf{Ancuti} & \textbf{Ours} \\
		\midrule
		\textbf{\textit{ancuti1}}  & 0.8366 &	1.8894 &  0.5132 & \textbf{2.7105} & 1.8259 & 0.5359
		\\
		\textbf{\textit{ancuti2}}  & 0.4490 &	\textbf{1.2412}	& 0.4546 & 0.9285 & 1.1200 &  0.5497
		\\
		\textbf{\textit{ancuti3}}  & 0.7487 &	0.9458 & 0.5766 & 1.0913 &  1.0264 &	\textbf{4.0781}
		\\
		\textbf{\textit{galdran}}  & 0.9045 &	4.7779 & 0.5776 & 1.4077 &  4.0371 &	\textbf{6.9439}
		\\
		\bottomrule
	\end{tabular}
	\label{tab:UCIQEResults}
\end{table}

\section{Conclusion} \label{sec:conc}

In this paper, we presented a methodology to restore the visual quality of underwater images. Our approach uses a set of image quality metrics to guide the learning process. It consists of a deep learning model that receives an underwater image and outputs a transmission map. This map is used to obtain a set of properties from the scene. The image formation model is then used to recover the underwater scene visual information. The experiments showed that our method improves the visual quality of underwater images by not degrading their edges and also performed well on the UCIQE metric.

\paragraph*{Acknowledgments.} \label{sec:ack}

The authors would like to thank CAPES, CNPq, and FAPEMIG for funding this work.

\bibliographystyle{IEEEbib}

\bibliography{refs}

\end{document}